\documentclass[11pt]{article}
\usepackage[utf8]{inputenc}
\usepackage[final]{acl}
\usepackage{multirow, makecell}
\usepackage{amsmath}
\usepackage{algorithm}
\usepackage{algpseudocode}
\usepackage{amssymb}

\usepackage{times}
\usepackage{latexsym}
\usepackage{graphicx}

\usepackage{xcolor,colortbl}
\usepackage[T1]{fontenc}
\usepackage[utf8]{inputenc}
\usepackage{tcolorbox}
\tcbuselibrary{breakable}
\usepackage{mwe}
\usepackage{tabu}
\usepackage{booktabs}
\usepackage{microtype}
\usepackage{subcaption} 

\title{NCL-UoR at SemEval-2025 Task 3: Detecting Multilingual Hallucination and Related Observable Overgeneration Text Spans with Modified RefChecker and Modified SeflCheckGPT}
\author{
Jiaying Hong\textsuperscript{1}, Thanet Markchom\textsuperscript{2}, Jianfei Xu\textsuperscript{1}, Tong Wu\textsuperscript{3} \and Huizhi Liang\textsuperscript{1} \\
\textsuperscript{1} School of Computing, Newcastle University, Newcastle upon Tyne, UK \\
\textsuperscript{2} Department of Computer Science, University of Reading, Reading, UK \\
\textsuperscript{3} Previously at School of Computing, Newcastle University, Newcastle upon Tyne, UK \\
\texttt{\small hongjalynn@gmail.com}, 
\texttt{\small thanet.markchom@reading.ac.uk}, \\
\texttt{\small mr.xujianfei@gmail.com}, 
\texttt{\small tongwuwhitney@gmail.com},
\texttt{\small huizhi.liang@newcastle.ac.uk}
}

\begin{document}

\newcommand{\methodone}{MRC}
\newcommand{\methodtwo}{MSCGH}

\newcommand{\identifierSelfGoogleF}{NCL-UoR\_Self\_GPT4o\_Google\_CSE}
\newcommand{\identifierSelfWiki}{NCL-UoR\_SelfModify-H-plus}
\newcommand{\identifierSelfGoogleA}{NCL-UOR\_CLAUDE-Modifier}

\newcommand{\approachSelfGoogleF}{MSCGH\_GPT\_CSE\_F}
\newcommand{\approachSelfWiki}{MSCGH\_GPT\_WIKI\_A}
\newcommand{\approachSelfGoogleA}{MRC\_CLAUDE\_CSE\_A}

\maketitle

\begin{abstract}
SemEval-2025 Task 3 (Mu-SHROOM) focuses on detecting hallucinations in content generated by various large language models (LLMs) across multiple languages. This task involves not only identifying the presence of hallucinations but also pinpointing their specific occurrences. To tackle this challenge, this study introduces two methods: Modified-RefChecker (\methodone) and Modified-SelfCheckGPT-H (\methodtwo). \methodone~integrates prompt-based factual verification into References, structuring them as claim-based tests rather than single external knowledge sources. \methodtwo~incorporates external knowledge to overcome its reliance on internal knowledge. In addition, both methods' original prompt designs are enhanced to identify hallucinated words within LLM-generated texts.
Experimental results demonstrate the effectiveness of the approach, achieving a high ranking on the test dataset in detecting hallucinations across various languages, with an average IoU of 0.5310 and an average COR of 0.5669. The source code used in this paper is available at \url{https://github.com/jianfeixu95/NCL-UoR}.
\end{abstract}

\section{Introduction}
Large language models (LLMs) have significantly advanced in producing human-like text across various domains~\cite{xiong-etal-2024-ncl,zhao-etal-2024-ncl}. However, one critical challenge remains: hallucinations—instances where the generated output contains logical inconsistencies, factual inaccuracies, or irrelevant information \cite{goodrich2019assessing}. These issues are particularly prominent in multilingual settings, where linguistic differences, cultural context, and the availability of external resources introduce additional complexities ~\cite{guerreiro-etal-2023-hallucinations}. To address this issue, SemEval-2025 Task-3: the Multilingual Shared-task on Hallucinations and Related Observable Overgeneration Mistakes (Mu-SHROOM)~\cite{vazquez-etal-2025-mu-shroom} was introduced. This task involves identifying hallucinated text spans in LLM-generated outputs across multiple languages and LLMs.  

To tackle this task, this work modifies two state-of-the-art methods: RefChecker \cite{hu2024refchecker} and SelfCheckGPT~\cite{manakul2023selfcheckgpt}. RefChecker detects fine-grained hallucinations by extracting claim triplets (subject, predicate, object) from LLM outputs and comparing them with pre-built reference data, using text classification and aggregation rules. However, it cannot precisely locate hallucination positions and relies on fixed and incomplete references. The proposed modified RefChecker improves upon this by introducing prompt-based fact verification, structuring references as claim-based tests for greater flexibility, and enhancing hallucination detection by calculating hallucination probabilities and providing soft and hard labels for more precise analysis.

SelfCheckGPT detects hallucinations by prompting the same LLM for multiple responses and identifying inconsistencies. However, reliance on internal knowledge may fail when hallucinations are consistent. To address this, we modify SelfCheckGPT by incorporating external knowledge and enhancing the prompt design to identify specific hallucinated words rather than only their presence.


Overall, unlike the original RefChecker and SelfCheckGPT, which rely on static references and internal prompt-based self-consistency, respectively, our modified methods incorporate external knowledge retrieval and prompt-driven span-level verification to improve hallucination detection accuracy and granularity.




\section{Related Work}

Most recent approaches to detecting hallucinations in LLM outputs rely on prompting techniques, where the models evaluate the likelihood of hallucinations in their responses. For instance, \citet{kadavath2022language} proposed prompting LLMs to generate an answer and then predict the probability of its correctness. 
\citet{manakul2023selfcheckgpt} introduced SelfCheckGPT, which compares an LLM-generated sentence against multiple alternative generations, asking the model to assess whether the original sentence is consistently supported. \citet{friel2023chainpoll} presented ChainPoll, using detailed prompts to guide models in identifying hallucinations.
\citet{hu2024refchecker} proposed RefChecker, a retrieval-augmented evaluation method that checks the consistency of model outputs against retrieved external references, aiming to identify factual inconsistencies and hallucinations without relying solely on LLM self-judgment. 
However, most existing methods focus on detecting whether a text contains hallucinations or not. Identifying the specific parts of a text that are hallucinations remains an open research challenge. Therefore, in this work, we modified RefChecker and SelfCheckGPT, two state-of-the-art methods to handle this task. 

\color{black}

\section{Methodology}

\begin{figure*}[h!]
    \centering
    \includegraphics[width=.85\linewidth,height=8.25cm]{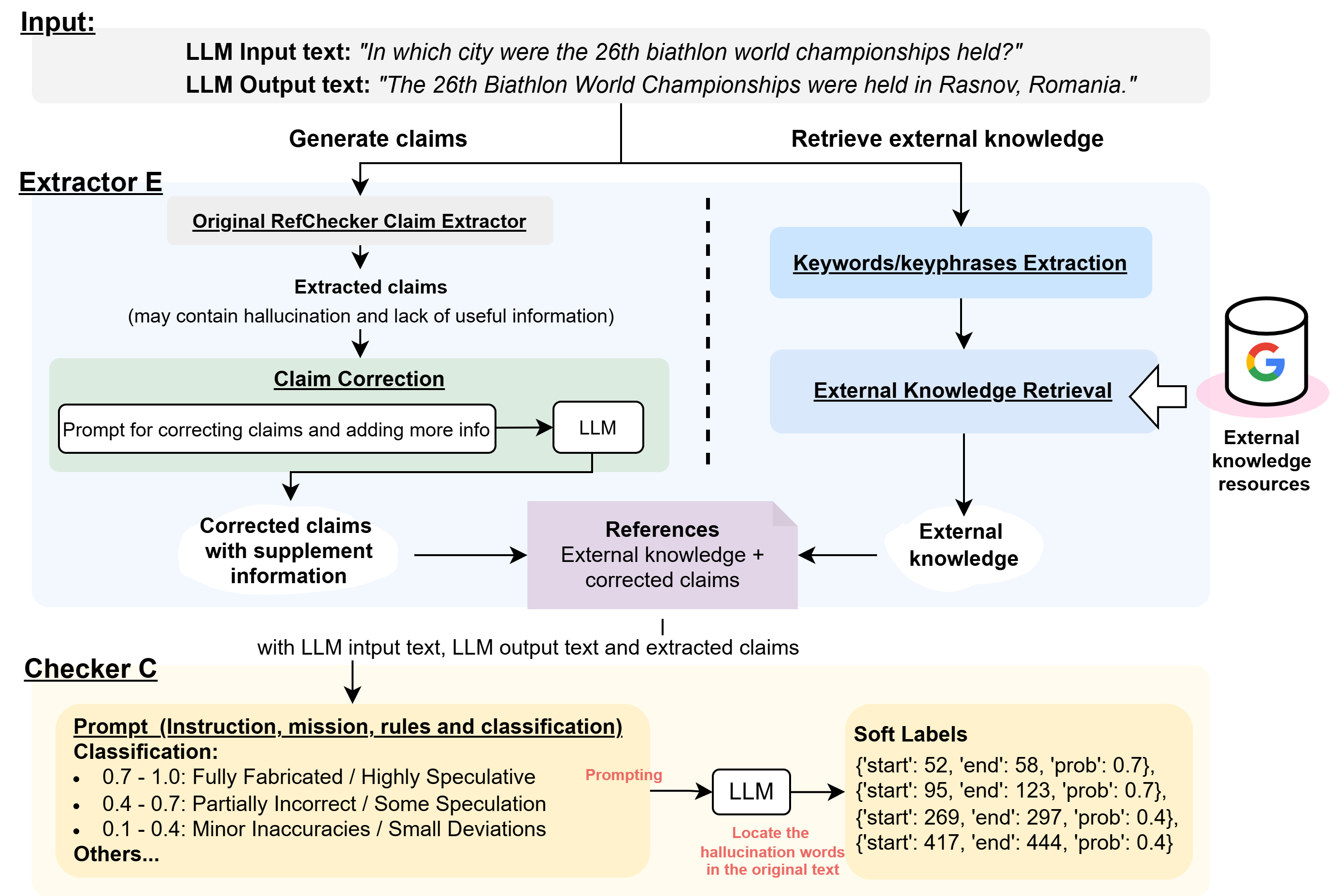}
    \caption{Overview of \methodone}
    \label{fig:diagram-CLAUDE-Modifier}
\end{figure*}

\subsection{Modified-RefChecker (\methodone)}
\methodone~ is an improved RefChecker, integrating CLAUDE \cite{anthropic_api} for enhanced functionality. Note that any LLM, including open-source ones, can be substituted. However, to ensure consistent and scalable evaluation, we adopt CLAUDE due to its multilingual support, API stability, and superior performance compared to open-source models in the original RefChecker \cite{hu2024refchecker}. \methodone~consists of two key components: the Extractor for constructing references and the Checker for identifying hallucinated words along with their probabilities. Figure \ref{fig:diagram-CLAUDE-Modifier} shows the overview of \methodone. The details of each component are described below.

\paragraph{Extractor Component} This component retrieves external knowledge using keywords or keyphrases through the Google CSE (Custom Search Engine) API \cite{EsraaQ.Naamha2023} (summarized search websites) and extracts claims from LLM responses, structured as triplets (subject, predicate, object), to form factual references. The extraction of claims utilizes the prompt design from RefChecker's Extractor \cite{hu2024refchecker} and is implemented using the Anthropic API \cite{anthropic_api}. However, the verification and refinement of claims are also conducted through the CLAUDE API, with the prompt design as in Appendix \ref{appendix:prompts} (Figure \ref{fig:prompt_CLAUDE_extractor}).

\paragraph{Checker Component} The Checker component evaluates hallucinated words and their probabilities in the model output by validating them against references using prompts. The prompts guide the classification of hallucinations and define their probabilities. The prompt design is as in Appendix \ref{appendix:prompts} (Figure \ref{fig:prompt_CLAUDE_checker}). With the support of CLAUDE API \cite{anthropic_api}, the results from Checker are mapped to the LLM output text, highlighting hallucinated words and generating soft labels and hard labels. Soft labels are based on the detected hallucination probabilities, while hard labels are determined by a threshold of 0.5 (probabilities > 0.5 are marked as hallucinations).

\subsection{Modified-SelfCheckGPT-H (\methodtwo)}
\label{sub:SelfModify-H}

\begin{figure*}[h]
    \centering
    \includegraphics[width=.95\linewidth,height=7.15cm]{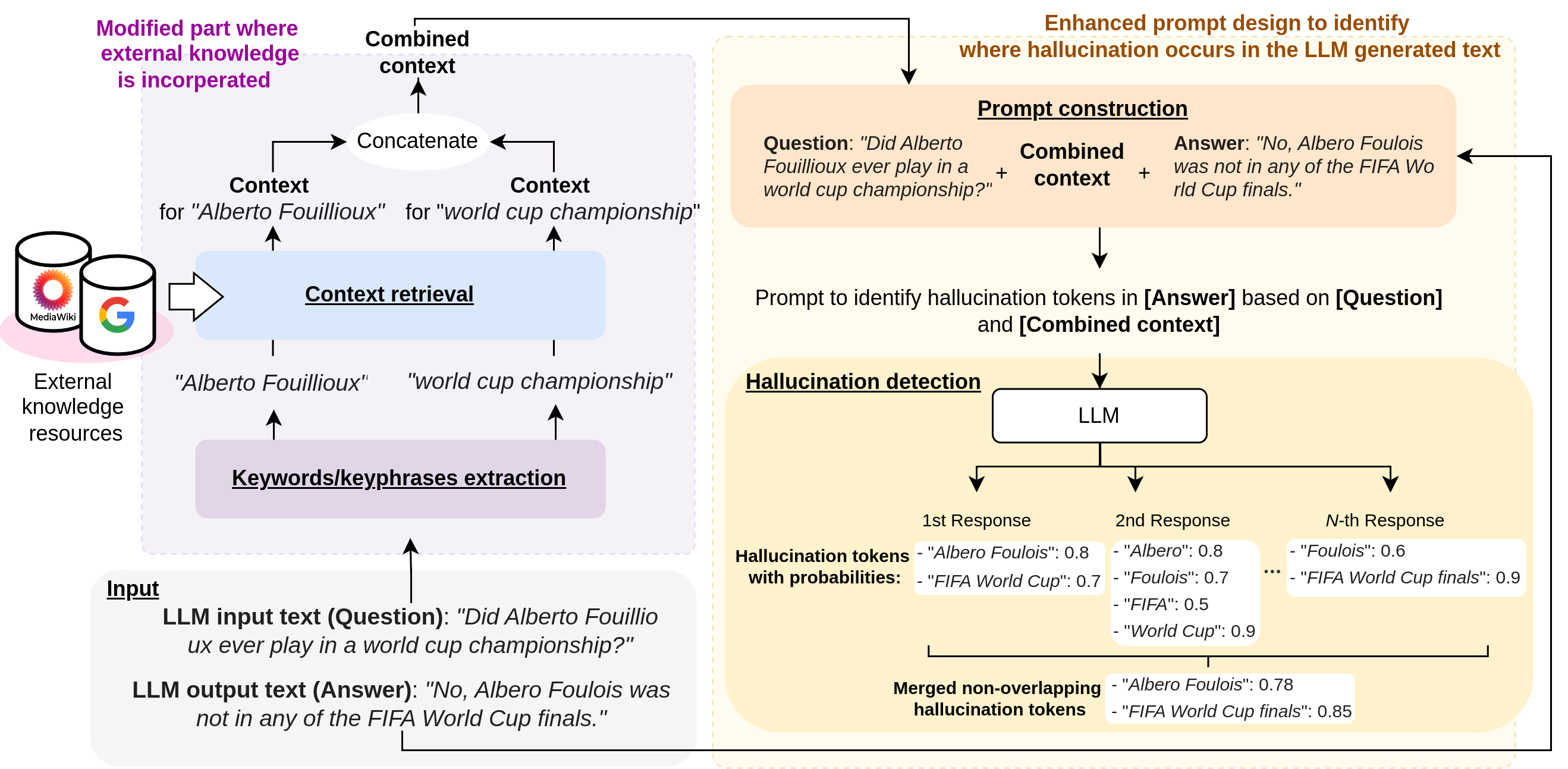}
    \caption{Overview of \methodtwo}
    \label{fig:ig:diagram-SelfModify-H}
\end{figure*}

\methodtwo~ is based on the method proposed by \citet{markchom-etal-2024-nu}. It consists of 4 steps: keywords/keyphrases extraction, context retrieval, prompt construction and hallucination detection. Figure \ref{fig:ig:diagram-SelfModify-H} shows an overview of \methodtwo. The details of each step are discussed in the following.

\paragraph{Keywords/Keyphrases Extraction} To generate a context for each LLM output text, keywords/keyphrases in the input text are first identified. In this work, YAKE (Yet Another Keyword Extractor)~\cite{CAMPOS2020257} is adopted to extract keywords across multiple languages, as it is domain- and language-independent. However, some languages are not covered by this method. Therefore, to improve keyword extraction in different languages, Hugging Face models are used for specific languages to identify named entities, while SpaCy facilitates tokenization and stop word removal. A summary of the tools and models used is shown in Appendix  \ref{appendix:custom_rules} (Table \ref{tab:tools_models}). Furthermore, GPT-3.5 was also applied to directly extract keywords/keyphrases from the LLM input text.

\paragraph{Context Retrieval} To retrieve a context based on each extracted keyword WikipediaAPI\footnote{https://pypi.org/project/Wikipedia-API/} \cite{wikipediaapi_online} and Google CSE API \cite{EsraaQ.Naamha2023} are considered. These resources are chosen for their popularity and capability to provide reliable context~\cite{Trokhymovych2021}. Once contexts for individual keywords are retrieved, they are concatenated to form a complete context for the LLM output text.

\paragraph{Prompt Construction}

Two prompt designs for identifying hallucinated words are explored. Prompt 1, adapted from SelfCheckGPT, is a simple prompt that asks an LLM to identify hallucinated words without specific instructions. Prompt 2, designed to identify hallucinated words, categorizes hallucination types and assigns probabilities while defining detection scope and output conditions. Compared to Prompt 1, it imposes stricter constraints to reduce unnecessary results \cite{rashkin2021faithfulness}. By directly classifying hallucinations and modeling probability distributions, it mitigates misalignment issues in LLM-generated text, improving detection accuracy and consistency.

\paragraph{Hallucination Detection}
    To detect hallucination words, an LLM (this work considers GPT-3.5, GPT-4 and GPT-4o following the original methodology of using GPT models in SelfCheckGPT \cite{manakul2023selfcheckgpt}.) is used to answer the prompt created in the previous step for each LLM output text. For each response, the hallucination words are identified, and a list of index intervals indicating the positions of these words in the LLM output string, $\mathcal{L} = \{L_1, L_2, \dots, L_n\}$, is obtained. Then, all overlapping and adjacent intervals across all $N$ responses are merged into a set of distinct, non-overlapping intervals $\mathcal{M} = \{(s_1, e_1), (s_2, e_2), \dots, (s_m, e_m)\}$. 

    The soft probabilities for each merged interval are computed differently depending on the prompt used (Prompt 1 or Prompt 2).
    For Prompt 1, the probability of each merged interval $(s_i, e_i)$ is computed by $p(s_i, e_i) = \frac{1}{n} \sum_{k=1}^{n} \frac{o_{i,k}}{e_i - s_i}$, where $o_{i,k}$ is the total overlap between the merged interval $(s_i, e_i)$ and the intervals in the list $L_k$ and $e_i - s_i$ is the length of the interval.   
   
    Prompt 2 detects the probabilities of hallucinated words. However, in the repeated \(N\) times process, the probabilities need to be recalculated, leading to the introduction of the following formula: $p(s_i, e_i) = \left( \frac{\sum_{k=1}^{n} o_{i,k} \cdot p_k}{\sum_{k=1}^{n} o_{i,k}} \right)^{1.2}$, where $p_k$ is the probability of hallucination for each interval in the $n$ responses, which is combined with the overlap length $o_{i,k}$ to calculate the weighted average probability. The exponent $1.2$ introduces non-linearity, giving higher importance to intervals with frequent overlaps and improving the accuracy of hallucination detection. All merged intervals in $\mathcal{M}$, along with their probabilities, serve as soft labels. Hard labels are obtained by selecting the intervals in $\mathcal{M}$ with probabilities higher than a predefined threshold, which is set to 0.5 in this work.

\section{Datasets and Experimental Setup}


\paragraph{Dataset} The datasets used in this study are provided by the organizers of Mu-SHROOM. The validation set, which contains annotated labels, was used for model development and tuning. In the final experiments, the test set was employed to comprehensively evaluate the performance of the models. The validation set includes data in 10 languages, along with LLM input texts, LLM-generated texts, LLM tokens, corresponding logit values, and hallucination annotations in the form of soft and hard labels, indicating both the locations and probabilities of hallucinations. The test set contains data in 14 languages. The models were evaluated independently for each language to ensure a comprehensive assessment across multilingual data.

\paragraph{Method Selection} Initially, evaluation was conducted on \methodone~ and five variations of \methodtwo~, each utilizing different keyword extractors, context retrieval tools, prompt designs, and LLMs for hallucination detection (as discussed in Section \ref{sub:SelfModify-H}). This resulted in six different models, each applied to 14 languages, for a total of 84 experiments. The details of these methods and their performance results can be found in Appendix \ref{appendix:all_results}. Each model is assigned a Submitted Identifier, which corresponds to the Identifier submitted on the official website\footnote{https://helsinki-nlp.github.io/shroom/}.
Based on the performance, the three best methods were selected for discussion:  
(1) \textbf{MRC\_CLAUDE\_CSE\_A}: MRC using GPT-3.5 for keyword extraction, Google CSE API (abstract only) for context retrieval, and CLAUDE for hallucination detection.  
(2) \textbf{MSCGH\_GPT\_CSE\_F}: MSCGH using GPT-3.5 for keyword extraction, full Google CSE API results, and GPT-4o for hallucination detection ($N=5$).  
(3) \textbf{MSCGH\_GPT\_WIKI\_A}: MSCGH using custom rules for keyword extraction, first 200 characters Wikipedia API results, and GPT-4o for hallucination detection ($N=5$).

\paragraph{Evaluation Metrics} The metrics provided by the organizers were used: \textbf{Intersection-over-Union (IoU) of Characters}: Measures the overlap between hallucinated characters marked in the gold reference and those predicted by the system, and \textbf{Probability Correlation}: Assesses how well the probability assigned by the system for a character being part of a hallucination correlates with the probabilities observed in human annotations. 

\paragraph{Baseline} Three baselines were provided in the task \cite{Vazquez2025MuSHROOM}: (1) Baseline (neural): Fine-tuning of the neural network classifier based on XLM-R, outputting binary (0/1) probability predictions for each token, (2) Baseline (mark-all): Predicting all characters as hallucinations ($probability = 1$), and (3) Baseline (mark-none): Predicting all characters as non-hallucinations ($probability = 0$).

\section{Results and Discussions}

\paragraph{Overall comparison of the proposed methods}

\begin{figure}[ht]
    \centering
    \begin{subfigure}{0.49\textwidth}
        \centering
        \includegraphics[width=\textwidth]{Task_3/img/performance_iou_baseline.png}
        \caption{Performance Across Languages (IoU Score)}
        \label{fig:performance_iou_baseline}
    \end{subfigure}
    
    \begin{subfigure}{0.49\textwidth}
        \centering
        \includegraphics[width=\textwidth]{Task_3/img/performance_cor_baseline.png}
        \caption{Performance Across Languages (COR Score)}
        \label{fig:performance_cor_baseline}
    \end{subfigure}
    
    \begin{subfigure}{0.45\textwidth}
        \centering
        \includegraphics[width=\textwidth]{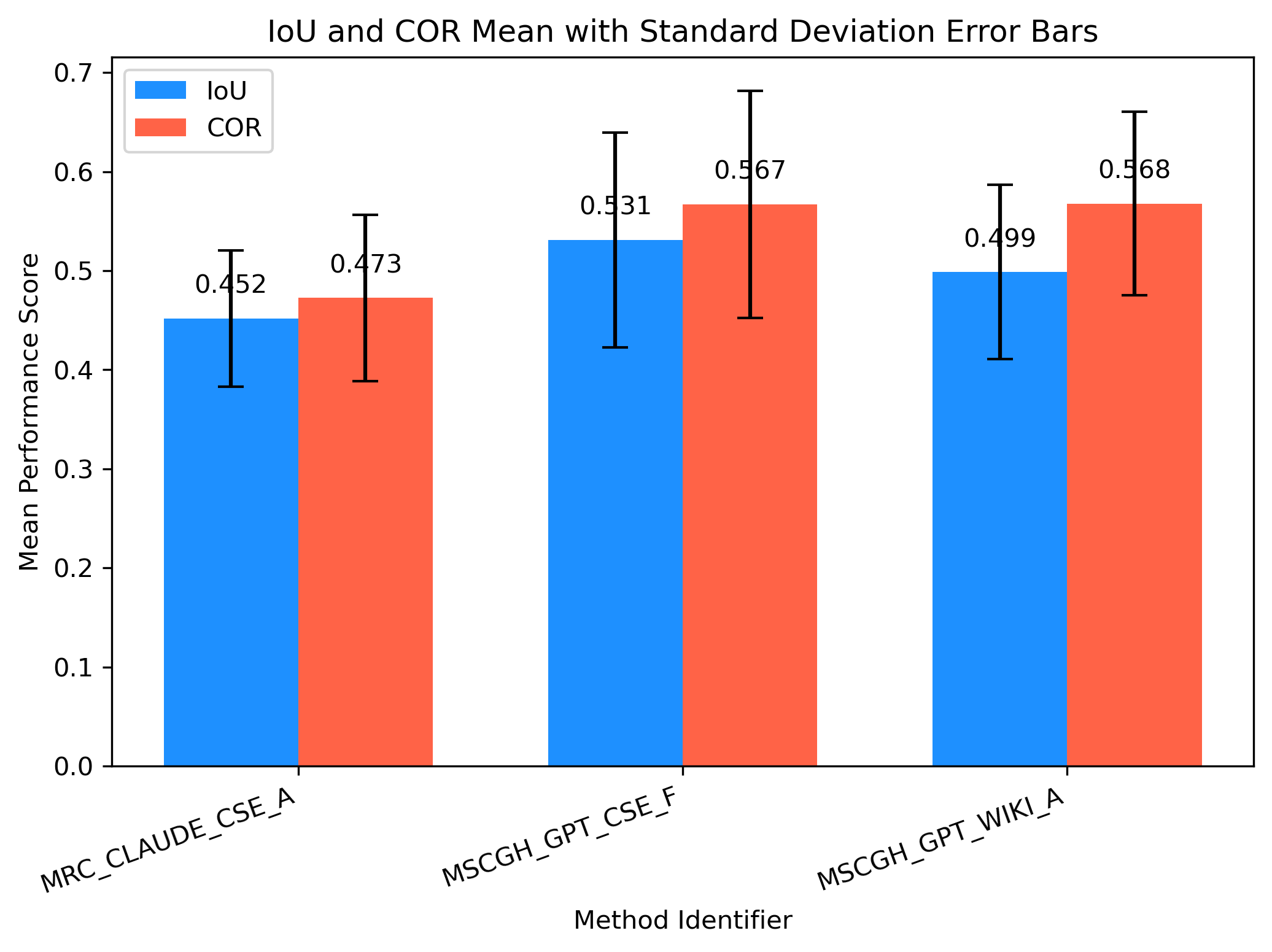}
        \caption{Methods Performance and Stability Comparison}
        \label{fig:methods_performance_and_stability}
    \end{subfigure}
    
    \caption{Performance Across Languages of Methods}
    \label{fig:perfomance_methods_languages}
\end{figure}

\begin{figure*}[htb]
    \centering
        \includegraphics[width=\textwidth]{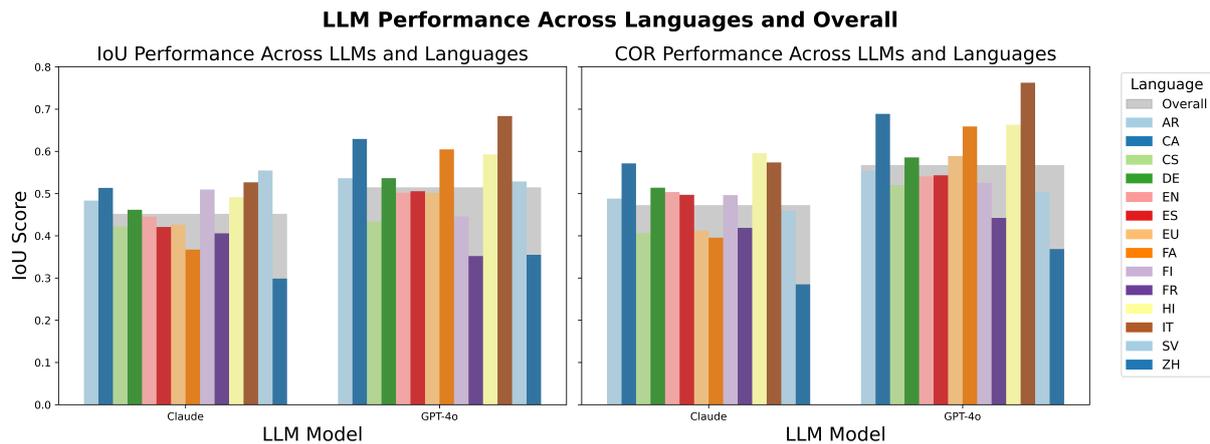}
        \caption{LLM Performance Across Languages and Overall}
        \label{fig:LLM_model_comparison}
\end{figure*}

The comparative results of our methods and the baselines are shown in Figures ~\ref{fig:performance_iou_baseline} and ~\ref{fig:performance_cor_baseline}. From these figures, the neural and mark-none baselines performed the worst across all languages, while the mark-all baseline achieved slightly higher IoU but nearly zero COR scores. In contrast, our method outperformed these baselines in all languages, with average improvements of approximately 0.30 in IoU and 0.45 in COR. More importantly, according to the 100,000 bootstrap resamplings mentioned in \cite{Vazquez2025MuSHROOM}, our submitted methods achieved a $Pr(rank)$ above 0.5 in every language. This indicates a higher probability of outperforming the next-best team in the majority of samples, and thus demonstrates robust and consistent cross-lingual performance.

Across multiple languages, our models exhibited distinct performance differences, as shown in Figure \ref{fig:perfomance_methods_languages}. \approachSelfGoogleF~ consistently led, benefiting from more effective keyword extraction, comprehensive external knowledge retrieval, and stronger hallucination detection. Its broader retrieval strategy provided an advantage in handling ambiguous or multi-step queries. \approachSelfWiki~ followed closely, particularly excelling in Chinese results, where complex segmentation and word relationships were better handled through its customized keyword extraction. \approachSelfGoogleA, while still effective, showed greater variance across languages, likely due to less optimized retrieval strategies or weaker hallucination detection.

Figure \ref{fig:methods_performance_and_stability} presents the average IoU and COR scores across all languages, illustrating the overall performance and stability of the three methods. \approachSelfGoogleF~ achieved the highest IoU and COR scores, while \approachSelfWiki~ performed similarly to \approachSelfGoogleA. However, MSCGH methods exhibited larger error bars, indicating greater variability and less stability. The fluctuations in MSCGH may stem from differences in knowledge retrieval and prompt design. In contrast, MRC demonstrated more consistent performance, suggesting its higher stability.


\paragraph{Comparison of knowledge retrieval methods}



In Figure \ref{fig:methods_performance_and_stability}, \approachSelfGoogleF~outperformed \approachSelfWiki. This could be attributed to differences in external knowledge resources and the precision of keyword extraction. GPT-3.5, as a keyword extraction tool, likely understood the context of questions better and extracted more precise and relevant keywords for retrieval. In contrast, custom rules had limitations in generalization and contextual understanding. They relied on specific language resources, which were limited in scope. This could affect the accuracy of keyword extraction and subsequently reduce the relevance and coverage of retrieved information. Besides keyword extraction, knowledge resources also played a vital role. The Google CSE API encompassed the Wikipedia API and extended beyond it, providing broader search coverage through a customizable search engine \cite{EsraaQ.Naamha2023}. Additionally, retrieving full-page content via the Google CSE API could yield better results than retrieving only abstract content, as suggested by \approachSelfGoogleF's superior performance over \approachSelfGoogleA. Overall, both keyword extraction accuracy and knowledge coverage influenced model performance. This highlights the importance of optimizing external knowledge extraction methods to improve detection outcomes.


\paragraph{Comparison of prompted LLMs for hallucination detection}

Figure \ref{fig:LLM_model_comparison} compares different LLMs. In this figure, each model is represented with bars showing the IoU and COR scores for individual languages. The grey bar behind each model's score bars indicates the average IoU and COR scores for that model. From this figure, CLAUDE performed the worst, while GPT-4o showed significant improvements. However, not all GPT-4o-based methods outperformed CLAUDE, indicating that LLM upgrades alone do not guarantee better results—effective knowledge retrieval remained essential. Performance gaps between LLMs were more pronounced in high-resource languages (e.g., Italian), where GPT-4o significantly outperformed CLAUDE. In contrast, for low-resource languages (e.g., Arabic), GPT-4o’s benefits were inconsistent—some methods showed only marginal gains, while those using the Google CSE API achieved substantial improvements. This underscored the critical role of external knowledge integration in maximizing LLM performance.


\section{Conclusion}


SemEval-2025 Mu-SHROOM introduced the task of detecting hallucination spans in multilingual LLM outputs. To tackle this task, this work proposed two methods: Modified-RefChecker (\methodone) and Modified-SelfCheckGPT-H (\methodtwo). These methods incorporated external knowledge integration and an improved prompt design, enabling the detection of text-span hallucinations in LLM-generated texts. \methodone~and variations of \methodtwo~ (with different keyword extraction techniques, external knowledge sources, and prompt strategies) were evaluated across datasets in 14 languages. Three top-performing methods were chosen for discussion in this paper. Among the evaluated methods, \methodtwo~using GPT-3.5 for keyword extraction, full Google CSE API results, and GPT-4o for hallucination detection achieved the best overall performance. Although \methodtwo~ demonstrated higher performance, it lacked stability when applied across different languages. Meanwhile, \methodone~was more stable but less optimized. 
One limitation of the proposed approaches is the assumption that external knowledge is accurate. However, the retrieved information may not always be fully factual due to the ever-growing volume of online content. Such inaccuracies could reduce the effectiveness of the proposed approaches.
Future research could focus on refining the prompt design and enhancing external knowledge integration and faulty correction strategies. Additionally, adaptive learning for low-resource languages and broader language task expansion could be considered.


\bibliography{task3-refs}

\appendix

\section{Prompts}
\label{appendix:prompts}

This appendix presents the prompts used in RefChecker (\methodone) and modified SelfCheckGPT (\methodtwo).
Figure \ref{fig:prompt_CLAUDE_extractor} shows the Claims Correction Prompt, used in \methodone. 
Figure \ref{fig:prompt_CLAUDE_checker} shows the Checker Component Prompt for \methodone. 
Figure \ref{fig:prompt_selfmodify_1} shows Prompt 1 for \methodtwo.
Figure \ref{fig:prompt_selfmodify_2} shows Prompt 2 for \methodtwo.

\begin{figure}[h!]
\centering
    \medskip  
        \noindent
        \scalebox{.7}{
        \fbox{%
        \centering
            \parbox{1.385\linewidth}{%
                \textbf{Prompt} \\ \vspace{-0.5em} 
                \hrule height 0.4pt \vspace{0.5em} 
                 System Task: Please expand, provide additional relevant factual information and verify about the following claim \\
                 Claims: \{claims\} \\
                 - If the claim is accurate, not hallucination and complete, return the original claim.  \\
                 - If the claim is inaccurate, partial, or lacking detail, return a corrected, more detailed, and comprehensive factual statement.  
            }%
        }%
        }
        \medskip
    \caption{Claims Correction Prompt for the Extractor Component of \methodone}
    \label{fig:prompt_CLAUDE_extractor}

\end{figure}

\begin{figure}[h!]

    \centering
    \medskip  
        \noindent
        \scalebox{.69}{
        \fbox{%
        \centering
            \parbox{1.385\linewidth}{%
                \textbf{Prompt} \\ \vspace{-0.5em} 
                \hrule height 0.4pt \vspace{0.5em} 
                 System Task: Evaluate the model output text for hallucinations by comparing it to the provided references, existing fact, claims, and question (model input). Identify any hallucinated or potentially inaccurate parts in the entire model output text. Highlight the hallucinated word and assign a probability of the hallucination word in the `model\_output\_text`. \\
                 LLM input text: \{LLM input text\}  \\
                 Claims: \{claims\}  \\
                 References: \{references\}  \\
                 LLM output text: \{LLM output text\}  \\
                 \textbf{Instructions} \\ 
                 1. Compare each claim with the provided references, question and existing fact (internal knowledge).  \\
                 2. If a claim cannot be fully supported by the references, identify the hallucinated words and mark it to `model output text`.  \\
                 3. Return character-level offsets and assign hallucination probabilities.  \\
                 4. If the claim is fully supported, hallucination should not to be labeled.  \\
                 5. Assign hallucination probabilities based on the following criteria:  \\
                    - 0.7 - 1.0: Fully fabricated or highly speculative content with no supporting evidence.  \\
                    - 0.4 - 0.7: Partially incorrect or speculative content, but some evidence supports parts of the claim.  \\
                    - 0.1 - 0.4: Minor inaccuracies, such as spelling errors, wrong formatting, or small factual deviations.  \\
                 6. Ensure that the hallucinated words do not overlap or repeat. If overlapping occurs, merge them or separate them appropriately.  \\
                 7. Ensure the words are shown in the `model output text`.  \\
                 8. Highlight text in `model output text` that could potentially be a hallucination even if not explicitly listed in the claims.  \\
                 9. Return all the hallucinated words or phrases and assign each a hallucination probability (between 0 and 1).  \\
                 10. Do not filter out hallucinations based on low probability. Return results for any potential hallucination.  \\
                 11. Do not include any explanations, summaries, or additional text. Return the JSON list directly.  \\
                 12. Ensure all potential hallucinations are listed, even those with probabilities as low as 0.1. 
            }%
        }%
        }
        \medskip 
    \caption{Prompt for the Checker Component in \methodone}
    \label{fig:prompt_CLAUDE_checker}

\end{figure}

\begin{figure}[h!]

\centering
    \medskip  
        \noindent
        \scalebox{.7}{
        \fbox{%
        \centering
            \parbox{1.385\linewidth}{%
                \textbf{Prompt 1} \\ \vspace{-0.5em} 
                \hrule height 0.4pt \vspace{0.5em} 
                 Context: \{combined context\} \\
                 Sentence: \{LLM output text\} \\
                 Which tokens in the sentence are not supported by the context above? \\
                 Provide the answer in the form of a list of hallucination tokens separated by '|' without accompanying texts.
            }%
        }%
        }
        \medskip 
    \caption{Prompt 1 for the Hallucinations Detection in \methodtwo}
    \label{fig:prompt_selfmodify_1}
\end{figure}

\begin{table*}[h!]
        \centering
        \caption{Tools and Models Utilized for Keyword Extraction}
        \label{tab:tools_models}
        \resizebox{\textwidth}{!}{
        \begin{tabular}{llll}
        \toprule        
        \textbf{Language} & \textbf{Stop Word Removal Tool}                 & \textbf{NER Model}                                           & \textbf{Additional Model/Approach} \\ \midrule
        Chinese (zh)      & jieba and HIT\_stopwords \cite{gitcode_63e0e}   & Hugging Face (`xlm-roberta-large-finetuned-conll03-english`) & TF-IDF (jieba.analyse)             \\ 
        Arabic (ar)       & `stopwords-ar.txt` \cite{alrefaie2019arabic}   & Hugging Face (`asafaya/bert-base-arabic`)                    & Tokenization (Hugging Face, TF-IDF)        \\ 
        Hindi (hi)        & Indic NLP Library (`indic\_tokenize`) \cite{indicnlp_tokenize}  & Hugging Face (`xlm-roberta-large-finetuned-conll03-english`) & Tokenization (Indic Tokenizer)     \\
        Basque (eu)       & Stopwords-iso (`stopwords-eu.txt`) \cite{stopwords-eu}           & Hugging Face (`xlm-roberta-large-finetuned-conll03-english`) & TF-IDF (spaCy `xx\_ent\_wiki\_sm`)    \\ 
        Czech (cs)        & StopwordsISO \cite{stopwordsiso}   & Stanza \cite{stanza}  & Tokenization (Stanza, TF-IDF)          \\ 
        Farsi (fa)        & Hazm \cite{hazm}  & Hugging Face (`bert-fa-base-uncased-ner-arman`)              & Tokenization (Stanza, TF-IDF)      \\ 
        Catalan (ca)      & spaCy (ca\_core\_news\_sm)                                          & Hugging Face (`projecte-aina/roberta-base-ca-v2-cased-ner`)  & TF-IDF (spaCy `ca\_core\_news\_sm`)            \\ 
        English (en)      & spaCy (en\_core\_web\_sm)         & Hugging Face (`xlm-roberta-large-finetuned-conll03-english`) & TF-IDF (spaCy `en\_core\_web\_sm`)    \\ 
        Spanish (es)      & spaCy (es\_core\_news\_sm)         & Hugging Face (`xlm-roberta-large-finetuned-conll03-english`) & TF-IDF (spaCy `es\_core\_news\_sm`)   \\ 
        French (fr)       & spaCy (fr\_core\_news\_sm)         & Hugging Face (`xlm-roberta-large-finetuned-conll03-english`) & TF-IDF (spaCy `fr\_core\_news\_sm`)   \\ 
        German (de)       & spaCy (de\_core\_news\_sm)         & Hugging Face (`xlm-roberta-large-finetuned-conll03-english`) & TF-IDF (spaCy `de\_core\_news\_sm`)   \\ 
        Italian (it)      & spaCy (it\_core\_news\_sm)         & Hugging Face (`xlm-roberta-large-finetuned-conll03-english`) & TF-IDF (spaCy `it\_core\_news\_sm`)   \\ 
        Finnish (fi)      & spaCy (fi\_core\_news\_sm)         & Hugging Face (`xlm-roberta-large-finetuned-conll03-english`) & TF-IDF (spaCy `fi\_core\_news\_sm`)   \\ 
        Swedish (sv)      & spaCy (sv\_core\_news\_sm)         & Hugging Face (`xlm-roberta-large-finetuned-conll03-english`) & TF-IDF (spaCy `sv\_core\_news\_sm`)    \\ \bottomrule
        \end{tabular}
        }
\end{table*}

\begin{figure}[h!]

     \medskip
        \noindent
        \scalebox{0.7}{
        \fbox{%
        \centering
            \parbox{1.385\linewidth}{%
                        \textbf{Prompt 2} \\ \vspace{-0.5em} 
                        \hrule height 0.4pt \vspace{0.5em} 
                        Language: \{language\} \\
                        Question: \{LLM input text\} \\
                        Sentence: \{LLM output text\}  \\
                        Context (if available): \{context\} \\
                        \textbf{Task} \\ 
                        You are an AI model output evaluation expert, responsible for detecting hallucinated words in model output and assigning accurate probability scores to each hallucination. \\
                        1. Identify hallucinated words or phrases in the model output based on the question and background knowledge. \\
                           - A word or phrase is considered a hallucination if it:  \\
                             - Contradicts the background knowledge.  \\
                             - Is unverifiable or fabricated.  \\
                             - Contains logical inconsistencies.  \\   
                         2. Assign a probability score to each hallucinated word or phrase according to the following criteria:  \\
                           - Probability > 0.7: Severe factual errors or contradictions.  \\
                           - Probability 0.5 - 0.7: Unverifiable or speculative content.  \\
                           - Probability 0.3 - 0.5: Minor inconsistencies or unverifiable details.  \\
                           - Probability 0.1 - 0.3: Minor inaccuracies or vague ambiguities.  \\
                           - Do not label words with probability $\leq$ 0.1 (i.e., verifiable facts).  \\
                        \textbf{Additional Instructions} \\ 
                        - Do not mark redundant or overly generic words (e.g., "the", "a", "and") as hallucinations unless they introduce factual errors.  \\
                        - Pay special attention to:  \\
                          - Numerical data (e.g., dates, quantities, percentages).  \\
                          - Named entities (e.g., people, organizations, locations).  \\
                          - Logical contradictions (e.g., self-contradictions within the text).  \\
                        - If background knowledge is absent, base your judgment solely on internal consistency.  
                    }%
                }%
            }
    \caption{Prompt 2 for the Hallucinations Detection in \methodtwo}
    \label{fig:prompt_selfmodify_2}
\end{figure}

\section{Custom Rules of Keywords/keyphrases Extraction in \methodtwo}
\label{appendix:custom_rules}
Table \ref{tab:tools_models} shows custom rules of keywords/keyphrases extraction across various languages in \methodtwo.

\section{All Results}
\label{appendix:all_results}
Table \ref{tab:methods_results} shows the results of all the methods on the 14 languages test set.
\begin{table*}[ht]
    \caption{All Methods Test Results}
    \label{tab:methods_results}
    \centering
    \small 
    \renewcommand{\arraystretch}{1.2} 
    \setlength{\tabcolsep}{4pt} 
    \resizebox{\textwidth}{!}{
    \begin{tabular}{clp{6cm}p{3.5cm}llccc}
        \toprule
        \bf Language & \bf Framework & \bf Submitted \bf Identifier & \bf Keywords Extraction & \bf External Knowledge & \bf LLM & \bf N & \bf  \bf IoU & \bf COR \\
        \midrule
        \multirow{6}{*}{AR} 
        & \methodtwo & NCL-UoR\_Self\_GPT3.5\_YAKE\_Wiki & YAKE & Wikipedia API & gpt-3.5-turbo & 5.0 & 0.2485 & 0.2154 \\
        & \methodone & NCL-UoR\_CLAUDE-Modifier & gpt-3.5-turbo & Google CSE API (abstract) & CLAUDE-3-5-haiku-20241022 & - & 0.4834 & 0.4881 \\
        & \methodtwo & NCL-UoR\_SelfModify-H & custom rules (Table \ref{tab:tools_models}) & Wikipedia API & gpt-3.5-turbo & 5.0 & 0.3752 & 0.3707 \\
        & \methodtwo & NCL-UoR\_SelfModify-H-plus & custom rules (Table \ref{tab:tools_models}) & Wikipedia API & gpt-4o & 5.0 & 0.5389 & 0.5710 \\
        & \methodtwo & NCL-UoR\_Self\_GPT4o\_Google\_CSE & gpt-3.5-turbo & Google CSE API & gpt-4o & 5.0 & 0.5334 & 0.5350 \\
        & \methodtwo & NCL-UoR\_Self\_GPT4\_GPT3.5\_Google\_CSE & gpt-3.5-turbo & Google CSE API (abstract) & gpt-4-turbo & 5.0 & 0.4353 & 0.4539 \\
        \midrule

        \multirow{6}{*}{CA} 
        & \methodtwo & NCL-UoR\_Self\_GPT3.5\_YAKE\_Wiki & YAKE & Wikipedia API & gpt-3.5-turbo & 5.0 & 0.0.3650 & 0.3778 \\
        & \methodone & NCL-UoR\_CLAUDE-Modifier & gpt-3.5-turbo & Google CSE API (abstract) & CLAUDE-3-5-haiku-20241022 & - & 0.5135 & 0.5714 \\
        & \methodtwo & NCL-UoR\_SelfModify-H & custom rules (Table \ref{tab:tools_models}) & Wikipedia API & gpt-3.5-turbo & 5.0 & 0.4849 & 0.5423 \\
        & \methodtwo & NCL-UoR\_SelfModify-H-plus & custom rules (Table \ref{tab:tools_models}) & Wikipedia API & gpt-4o & 5.0 & 0.5984 & 0.6573 \\
        & \methodtwo & NCL-UoR\_Self\_GPT4o\_Google\_CSE & gpt-3.5-turbo & Google CSE API & gpt-4o & 5.0 & 0.6602 & 0.7202 \\
        & \methodtwo & NCL-UoR\_Self\_GPT4\_GPT3.5\_Google\_CSE & gpt-3.5-turbo & Google CSE API (abstract) & gpt-4-turbo & 5.0 & 0.4621 & 0.6072 \\
        \midrule

        \multirow{6}{*}{CS} 
        & \methodtwo & NCL-UoR\_Self\_GPT3.5\_YAKE\_Wiki & YAKE & Wikipedia API & gpt-3.5-turbo & 5.0 & 0.2121 & 0.2364 \\
        & \methodone & NCL-UoR\_CLAUDE-Modifier & gpt-3.5-turbo & Google CSE API (abstract) & CLAUDE-3-5-haiku-20241022 & - & 0.4218 & 0.4061 \\
        & \methodtwo & NCL-UoR\_SelfModify-H & custom rules (Table \ref{tab:tools_models}) & Wikipedia API & gpt-3.5-turbo & 5.0 & 0.2513 & 0.3189 \\
        & \methodtwo & NCL-UoR\_SelfModify-H-plus & custom rules (Table \ref{tab:tools_models}) & Wikipedia API & gpt-4o & 5.0 & 0.4409 & 0.5285 \\
        & \methodtwo & NCL-UoR\_Self\_GPT4o\_Google\_CSE & gpt-3.5-turbo & Google CSE API & gpt-4o & 5.0 & 0.4264 & 0.5110 \\
        & \methodtwo & NCL-UoR\_Self\_GPT4\_GPT3.5\_Google\_CSE & gpt-3.5-turbo & Google CSE API (abstract) & gpt-4-turbo & 5.0 & 0.3935 & 0.4816 \\
        \midrule

        \multirow{6}{*}{DE} 
        & \methodtwo & NCL-UoR\_Self\_GPT3.5\_YAKE\_Wiki & YAKE & Wikipedia API & gpt-3.5-turbo & 5.0 & 0.3295 & 0.3713 \\
        & \methodone & NCL-UoR\_CLAUDE-Modifier & gpt-3.5-turbo & Google CSE API (abstract) & CLAUDE-3-5-haiku-20241022 & - & 0.4617 & 0.5139 \\
        & \methodtwo & NCL-UoR\_SelfModify-H & custom rules (Table \ref{tab:tools_models}) & Wikipedia API & gpt-3.5-turbo & 5.0 & 0.4173 & 0.4601 \\
        & \methodtwo & NCL-UoR\_SelfModify-H-plus & custom rules (Table \ref{tab:tools_models}) & Wikipedia API & gpt-4o & 5.0 & 0.5259 & 0.5852 \\
        & \methodtwo & NCL-UoR\_Self\_GPT4o\_Google\_CSE & gpt-3.5-turbo & Google CSE API & gpt-4o & 5.0 & 0.5472 & 0.5860 \\
        & \methodtwo & NCL-UoR\_Self\_GPT4\_GPT3.5\_Google\_CSE & gpt-3.5-turbo & Google CSE API (abstract) & gpt-4-turbo & 5.0 & 0.4467 & 0.5001 \\
        \midrule

        \multirow{6}{*}{EN} 
        & \methodtwo & NCL-UoR\_Self\_GPT3.5\_YAKE\_Wiki & YAKE & Wikipedia API & gpt-3.5-turbo & 5.0 & 0.4245 & 0.4544 \\
        & \methodone & NCL-UoR\_CLAUDE-Modifier & gpt-3.5-turbo & Google CSE API (abstract) & CLAUDE-3-5-haiku-20241022 & - & 0.4451 & 0.5035 \\
        & \methodtwo & NCL-UoR\_SelfModify-H & custom rules (Table \ref{tab:tools_models}) & Wikipedia API & gpt-3.5-turbo & 5.0 & 0.3690 & 0.3905 \\
        & \methodtwo & NCL-UoR\_SelfModify-H-plus & custom rules (Table \ref{tab:tools_models}) & Wikipedia API & gpt-4o & 5.0 & 0.4844 & 0.5333 \\
        & \methodtwo & NCL-UoR\_Self\_GPT4o\_Google\_CSE & gpt-3.5-turbo & Google CSE API & gpt-4o & 5.0 & 0.5195 & 0.5476 \\
        & \methodtwo & NCL-UoR\_Self\_GPT4\_GPT3.5\_Google\_CSE & gpt-3.5-turbo & Google CSE API (abstract) & gpt-4-turbo & 5.0 & 0.4469 & 0.4690 \\
        \midrule

        \multirow{6}{*}{ES} 
        & \methodtwo & NCL-UoR\_Self\_GPT3.5\_YAKE\_Wiki & YAKE & Wikipedia API & gpt-3.5-turbo & 5.0 & 0.3129 & 0.3122 \\
        & \methodone & NCL-UoR\_CLAUDE-Modifier & gpt-3.5-turbo & Google CSE API (abstract) & CLAUDE-3-5-haiku-20241022 & - & 0.4206 & 0.4970 \\
        & \methodtwo & NCL-UoR\_SelfModify-H & custom rules (Table \ref{tab:tools_models}) & Wikipedia API & gpt-3.5-turbo & 5.0 & 0.3843 & 0.4104 \\
        & \methodtwo & NCL-UoR\_SelfModify-H-plus & custom rules (Table \ref{tab:tools_models}) & Wikipedia API & gpt-4o & 5.0 & 0.4964 & 0.5402 \\
        & \methodtwo & NCL-UoR\_Self\_GPT4o\_Google\_CSE & gpt-3.5-turbo & Google CSE API & gpt-4o & 5.0 & 0.5146 & 0.5464 \\
        & \methodtwo & NCL-UoR\_Self\_GPT4\_GPT3.5\_Google\_CSE & gpt-3.5-turbo & Google CSE API (abstract) & gpt-4-turbo & 5.0 & 0.4240 & 0.4790 \\
        \midrule

        \multirow{6}{*}{EU} 
        & \methodtwo & NCL-UoR\_Self\_GPT3.5\_YAKE\_Wiki & YAKE & Wikipedia API & gpt-3.5-turbo & 5.0 & 0.3111 & 0.2833 \\
        & \methodone & NCL-UoR\_CLAUDE-Modifier & gpt-3.5-turbo & Google CSE API (abstract) & CLAUDE-3-5-haiku-20241022 & - & 0.4263 & 0.4123 \\
        & \methodtwo & NCL-UoR\_SelfModify-H & custom rules (Table \ref{tab:tools_models}) & Wikipedia API & gpt-3.5-turbo & 5.0 & 0.4340 & 0.4907 \\
        & \methodtwo & NCL-UoR\_SelfModify-H-plus & custom rules (Table \ref{tab:tools_models}) & Wikipedia API & gpt-4o & 5.0 & 0.5104 & 0.5974 \\
        & \methodtwo & NCL-UoR\_Self\_GPT4o\_Google\_CSE & gpt-3.5-turbo & Google CSE API & gpt-4o & 5.0 & 0.4928 & 0.5802 \\
        & \methodtwo & NCL-UoR\_Self\_GPT4\_GPT3.5\_Google\_CSE & gpt-3.5-turbo & Google CSE API (abstract) & gpt-4-turbo & 5.0 & 0.3922 & 0.4932 \\
        \midrule

        \multirow{6}{*}{FA} 
        & \methodtwo & NCL-UoR\_Self\_GPT3.5\_YAKE\_Wiki & YAKE & Wikipedia API & gpt-3.5-turbo & 5.0 & 0.3254 & 0.3421 \\
        & \methodone & NCL-UoR\_CLAUDE-Modifier & gpt-3.5-turbo & Google CSE API (abstract) & CLAUDE-3-5-haiku-20241022 & - & 0.3672 & 0.3955 \\
        & \methodtwo & NCL-UoR\_SelfModify-H & custom rules (Table \ref{tab:tools_models}) & Wikipedia API & gpt-3.5-turbo & 5.0 & 0.5027 & 0.5653 \\
        & \methodtwo & NCL-UoR\_SelfModify-H-plus & custom rules (Table \ref{tab:tools_models}) & Wikipedia API & gpt-4o & 5.0 & 0.5509 & 0.6444 \\
        & \methodtwo & NCL-UoR\_Self\_GPT4o\_Google\_CSE & gpt-3.5-turbo & Google CSE API & gpt-4o & 5.0 & 0.6585 & 0.6732 \\
        & \methodtwo & NCL-UoR\_Self\_GPT4\_GPT3.5\_Google\_CSE & gpt-3.5-turbo & Google CSE API (abstract) & gpt-4-turbo & 5.0 & 0.4034 & 0.5500 \\
        \midrule

        \multirow{6}{*}{FI} 
        & \methodtwo & NCL-UoR\_Self\_GPT3.5\_YAKE\_Wiki & YAKE & Wikipedia API & gpt-3.5-turbo & 5.0 & 0.2983 & 0.3114 \\
        & \methodone & NCL-UoR\_CLAUDE-Modifier & gpt-3.5-turbo & Google CSE API (abstract) & CLAUDE-3-5-haiku-20241022 & - & 0.5095 & 0.4964 \\
        & \methodtwo & NCL-UoR\_SelfModify-H & custom rules (Table \ref{tab:tools_models}) & Wikipedia API & gpt-3.5-turbo & 5.0 & 0.3187 & 0.3656 \\
        & \methodtwo & NCL-UoR\_SelfModify-H-plus & custom rules (Table \ref{tab:tools_models}) & Wikipedia API & gpt-4o & 5.0 & 0.3928 & 0.4982 \\
        & \methodtwo & NCL-UoR\_Self\_GPT4o\_Google\_CSE & gpt-3.5-turbo & Google CSE API & gpt-4o & 5.0 & 0.4982 & 0.5523 \\
        & \methodtwo & NCL-UoR\_Self\_GPT4\_GPT3.5\_Google\_CSE & gpt-3.5-turbo & Google CSE API (abstract) & gpt-4-turbo & 5.0 & 0.3866 & 0.4906 \\
        \midrule

        \multirow{6}{*}{FR} 
        & \methodtwo & NCL-UoR\_Self\_GPT3.5\_YAKE\_Wiki & YAKE & Wikipedia API & gpt-3.5-turbo & 5.0 & 0.2094 & 0.2065 \\
        & \methodone & NCL-UoR\_CLAUDE-Modifier & gpt-3.5-turbo & Google CSE API (abstract) & CLAUDE-3-5-haiku-20241022 & - & 0.4058 & 0.4187 \\
        & \methodtwo & NCL-UoR\_SelfModify-H & custom rules (Table \ref{tab:tools_models}) & Wikipedia API & gpt-3.5-turbo & 5.0 & 0.3202 & 0.3685 \\
        & \methodtwo & NCL-UoR\_SelfModify-H-plus & custom rules (Table \ref{tab:tools_models}) & Wikipedia API & gpt-4o & 5.0 & 0.3571 & 0.4822 \\
        & \methodtwo & NCL-UoR\_Self\_GPT4o\_Google\_CSE & gpt-3.5-turbo & Google CSE API & gpt-4o & 5.0 & 0.3466 & 0.4024 \\
        & \methodtwo & NCL-UoR\_Self\_GPT4\_GPT3.5\_Google\_CSE & gpt-3.5-turbo & Google CSE API (abstract) & gpt-4-turbo & 5.0 & 0.3386 & 0.4712 \\
        \midrule

        \multirow{6}{*}{HI} 
        & \methodtwo & NCL-UoR\_Self\_GPT3.5\_YAKE\_Wiki & YAKE & Wikipedia API & gpt-3.5-turbo & 5.0 & 0.2251 & 0.1705 \\
        & \methodone & NCL-UoR\_CLAUDE-Modifier & gpt-3.5-turbo & Google CSE API (abstract) & CLAUDE-3-5-haiku-20241022 & - & 0.4914 & 0.5958 \\
        & \methodtwo & NCL-UoR\_SelfModify-H & custom rules (Table \ref{tab:tools_models}) & Wikipedia API & gpt-3.5-turbo & 5.0 & 0.5606 & 0.6078 \\
        & \methodtwo & NCL-UoR\_SelfModify-H-plus & custom rules (Table \ref{tab:tools_models}) & Wikipedia API & gpt-4o & 5.0 & 0.5570 & 0.6433 \\
        & \methodtwo & NCL-UoR\_Self\_GPT4o\_Google\_CSE & gpt-3.5-turbo & Google CSE API & gpt-4o & 5.0 & 0.6286 & 0.6830 \\
        & \methodtwo & NCL-UoR\_Self\_GPT4\_GPT3.5\_Google\_CSE & gpt-3.5-turbo & Google CSE API (abstract) & gpt-4-turbo & 5.0 & 0.5886 & 0.6664 \\
        \midrule

        \multirow{6}{*}{IT} 
        & \methodtwo & NCL-UoR\_Self\_GPT3.5\_YAKE\_Wiki & YAKE & Wikipedia API & gpt-3.5-turbo & 5.0 & 0.4153 & 0.4123 \\
        & \methodone & NCL-UoR\_CLAUDE-Modifier & gpt-3.5-turbo & Google CSE API (abstract) & CLAUDE-3-5-haiku-20241022 & - & 0.5265 & 0.5737 \\
        & \methodtwo & NCL-UoR\_SelfModify-H & custom rules (Table \ref{tab:tools_models}) & Wikipedia API & gpt-3.5-turbo & 5.0 & 0.6563 & 0.6941 \\
        & \methodtwo & NCL-UoR\_SelfModify-H-plus & custom rules (Table \ref{tab:tools_models}) & Wikipedia API & gpt-4o & 5.0 & 0.6547 & 0.7637 \\
        & \methodtwo & NCL-UoR\_Self\_GPT4o\_Google\_CSE & gpt-3.5-turbo & Google CSE API & gpt-4o & 5.0 & 0.7122 & 0.7613 \\
        & \methodtwo & NCL-UoR\_Self\_GPT4\_GPT3.5\_Google\_CSE & gpt-3.5-turbo & Google CSE API (abstract) & gpt-4-turbo & 5.0 & 0.5950 & 0.7313 \\
        \midrule

        \multirow{6}{*}{SV} 
        & \methodtwo & NCL-UoR\_Self\_GPT3.5\_YAKE\_Wiki & YAKE & Wikipedia API & gpt-3.5-turbo & 5.0 & 0.3763 & 0.2863 \\
        & \methodone & NCL-UoR\_CLAUDE-Modifier & gpt-3.5-turbo & Google CSE API (abstract) & CLAUDE-3-5-haiku-20241022 & - & 0.5546 & 0.4587 \\
        & \methodtwo & NCL-UoR\_SelfModify-H & custom rules (Table \ref{tab:tools_models}) & Wikipedia API & gpt-3.5-turbo & 5.0 & 0.4047 & 0.4335 \\
        & \methodtwo & NCL-UoR\_SelfModify-H-plus & custom rules (Table \ref{tab:tools_models}) & Wikipedia API & gpt-4o & 5.0 & 0.5233 & 0.5224 \\
        & \methodtwo & NCL-UoR\_Self\_GPT4o\_Google\_CSE & gpt-3.5-turbo & Google CSE API & gpt-4o & 5.0 & 0.5340 & 0.4836 \\
        & \methodtwo & NCL-UoR\_Self\_GPT4\_GPT3.5\_Google\_CSE & gpt-3.5-turbo & Google CSE API (abstract) & gpt-4-turbo & 5.0 & 0.4918 & 0.4907 \\
        \midrule

        \multirow{6}{*}{ZH} 
        & \methodtwo & NCL-UoR\_Self\_GPT3.5\_YAKE\_Wiki & YAKE & Wikipedia API & gpt-3.5-turbo & 5.0 & 0.1683 & 0.2840 \\
        & \methodone & NCL-UoR\_CLAUDE-Modifier & gpt-3.5-turbo & Google CSE API (abstract) & CLAUDE-3-5-haiku-20241022 & - & 0.2986 & 0.2849 \\
        & \methodtwo & NCL-UoR\_SelfModify-H & custom rules (Table \ref{tab:tools_models}) & Wikipedia API & gpt-3.5-turbo & 5.0 & 0.1849 & 0.2271 \\
        & \methodtwo & NCL-UoR\_SelfModify-H-plus & custom rules (Table \ref{tab:tools_models}) & Wikipedia API & gpt-4o & 5.0 & 0.3492 & 0.3830 \\
        & \methodtwo & NCL-UoR\_Self\_GPT4o\_Google\_CSE & gpt-3.5-turbo & Google CSE API & gpt-4o & 5.0 & 0.3606 & 0.3539 \\
        & \methodtwo & NCL-UoR\_Self\_GPT4\_GPT3.5\_Google\_CSE & gpt-3.5-turbo & Google CSE API (abstract) & gpt-4-turbo & 5.0 & 0.2842 & 0.3073 \\
        \bottomrule
    \end{tabular}
    }

\end{table*}

\end{document}